\documentclass[10pt,letterpaper]{article}

\usepackage{cogsci}

\cogscifinalcopy 

\usepackage{euler}
\usepackage{graphicx}
\usepackage{amsmath}
\usepackage{cleveref}
\newcommand\given[1][]{\:#1\vert\:}
\usepackage{xfrac}
\usepackage{url}
\usepackage{mathtools}
\usepackage{amsfonts}

\usepackage{pslatex}
\usepackage{apacite}
\usepackage{float} 

\title{Recovering Quantitative Models of Human Information Processing \\with Differentiable Architecture Search}

\author{{\large \bf Sebastian Musslick (musslick@princeton.edu)} \\
  Princeton Neuroscience Institute, Princeton University \\
  Princeton, NJ 08544, USA}

\begin{document}

\maketitle

\begin{abstract}
The integration of behavioral phenomena into mechanistic models of cognitive function is a fundamental staple of cognitive science. Yet, researchers are beginning to accumulate increasing amounts of data without having the temporal or monetary resources to integrate these data into scientific theories. We seek to overcome these limitations by incorporating existing machine learning techniques into an open-source pipeline for the automated construction of quantitative models. This pipeline leverages the use of neural architecture search to automate the discovery of interpretable model architectures, and automatic differentiation to automate the fitting of model parameters to data. We evaluate the utility of these methods based on their ability to recover quantitative models of human information processing from synthetic data.  We find that these methods are capable of recovering basic quantitative motifs from models of psychophysics, learning and decision making. We also highlight weaknesses of this framework and discuss future directions for their mitigation.

\end{abstract}

\section{Introduction}


The process of developing a mechanistic model of cognition incurs two challenges: (1) identifying the architecture of the model, i.e. the composition of functions and parameters, and (2) tuning parameters of the model to fit experimental data. While there are various methods for automating the fitting of parameters, cognitive scientists typically leverage their own expertise and intuitions to identify the architecture of a model---a process that requires substantial human effort. In machine learning, interest has grown in automating the construction and parameterization of neural networks to solve machine learning problems more efficiently \cite{he2021automl}. This involves the use of neural architecture search (NAS) for automating the discovery of model architectures \cite{elsken2019neural}, and the use of automatic differentiation to automate parameter fitting \cite{paszke2017automatic}. This combination of methods has led to breakthroughs in the automated construction of neural networks that are capable of outperforming networks designed by human researchers \cite<e.g. in computer vision:>{mendoza2016towards}. In this study, we explore the utility of these methods for constructing quantitative models of human information processing.


To ease the application of NAS to the discovery of a quantitative model, it is useful to treat quantitative models as neural networks or, more generally, as computation graphs. In this article, we introduce the notion of a computation graph and describe ways of expressing the architecture of a quantitative model in terms of a such a graph. We then review the use of differentiable architecture search \cite<DARTS;>{liu2018darts} for searching the space candidate computation graphs, and introduce an adaptation of this method for discovering quantitative models of human information processing.  We evaluate two variants of DARTS---regular DARTS \cite{liu2018darts} and fair DARTS \cite{chu2020fair}---based on their ability to recover three different models of human cognition from synthetic data, to explain behavioral phenomena in psychophysics, learning and perceptual decision making. Our results indicate that such algorithms are capable of recovering computational motifs found in these models. However, we also discuss further developments that are needed to expand the scope of models amenable to DARTS. Reported simulations (\url{https://github.com/musslick/DARTS-Cognitive-Modeling}) are embedded in a documented open-source framework for autonomous empirical research (\url{www.empiricalresearch.ai}) and can be extended to explore other search methods. 





\section{Quantitative Models as Computation Graphs}

A broad class of complex mathematical functions---including the functions expressed by a quantitative model of human information processing---can be formulated as a computation graph. A computation graph is a collection of nodes that are connected by directed edges. Each node denotes an expression of a variable, and each outgoing edge corresponds to a function applied to this variable (cf. Figure \ref{fig:compgraph}D). The value of a node is typically computed by integrating over the result of every function (edge) feeding to that node. Akin to quantitative models of cognitive function, a computation graph can take experiment parameters as input (e.g. the brightness of two visual stimuli), and can transform this input through a combination of functions (edges) and latent variables (intermediate nodes) to produce observable dependent measures as output nodes (e.g the probability that a participant is able to detect the difference in brightness between two stimuli). 

The expression of a formula as computation graph can be illustrated with Weber's law \cite{fechner1860elemente}---a quantitative hypothesis that relates the difference between the intensities of two stimuli to the probability that a participant can detect this difference. It states that the just noticeable difference (JND; the difference in intensity that a participant is capable of detecting in 50\% of the trials) amounts to

\vspace{-0.3cm}
\begin{equation}
\label{eqn:weber_law}
\Delta I = c \cdot I_0
\end{equation}

where $\Delta I$ is the JND, $I_0$ corresponds to the intensity of the baseline stimulus and $c$ is a constant. The probability of detecting the difference between two stimuli, $I_0$ and $I_1$, can then be formulated as a function of the two stimuli (with $I_0 < I_1$),

\vspace{-0.3cm}
\begin{equation}
\label{eqn:weber_model}
P(\textrm{detected}) = \sigma_\textrm{logistic}((I_1 - I_0) - \Delta I)
\end{equation}


where $\sigma_\textrm{logistic}$ is a logistic function. Figure \ref{fig:compgraph}D depicts the argument of $\sigma_\textrm{logistic}$  as a computation graph for $c=0.5$. The graph encompasses two input nodes, one representing $x_0=I_0$ and the other $x_1=I_1$. The intermediate node $x_2$ expresses $\Delta I$ which results from multiplying $I_0$ with the parameter $c=0.5$. The addition and subtraction of $I_1$ and $I_0$, respectively, result in their difference $(I_1 - I_0)$ and are represented by the intermediate node $x_3$. The linear combination of $x_2$ and $x_3$ in the output node $r$ resembles the argument to $\sigma_\textrm{logistic}$.

The automated construction of a mathematical hypothesis, like Weber's law, can be formulated as a search over the space of all possible computation graphs. Machine learning researchers leverage the notion of computation graphs to represent the composition of functions performed by a complex artificial neural network (i.e. its architecture), and deploy NAS to search a space of computation graphs. Although some level of specification of the graph remains with the researcher, NAS relieves the researcher from searching through these possibilities.


\section{Identifying Computation Graphs\\ with Neural Architecture Search}

NAS refers to a family of methods for automating the discovery of useful neural network architectures. There are a number of methods to guide this search, such as evolutionary algorithms, reinforcement learning or Bayesian optimization \cite<for a recent survey of NAS search strategies, see>{elsken2019neural}. However, most of these methods are computationally demanding due to the nature of the optimization problem: The search space of candidate computation graphs is high-dimensional and discrete. To address this problem, \citeA{liu2018darts}  proposed DARTS which relaxes the search space to become continuous, making architecture search amenable to gradient decent. The authors demonstrate that DARTS can yield useful network architectures for image classification and language modeling that are on par with architectures designed by human researchers. In this work, we assess whether variants of DARTS can be adopted to automate the discovery of interpretable quantitative models to explain human information processing.



\subsection{Regular DARTS}

\begin{figure}
\begin{center}
\includegraphics[scale = 0.35]{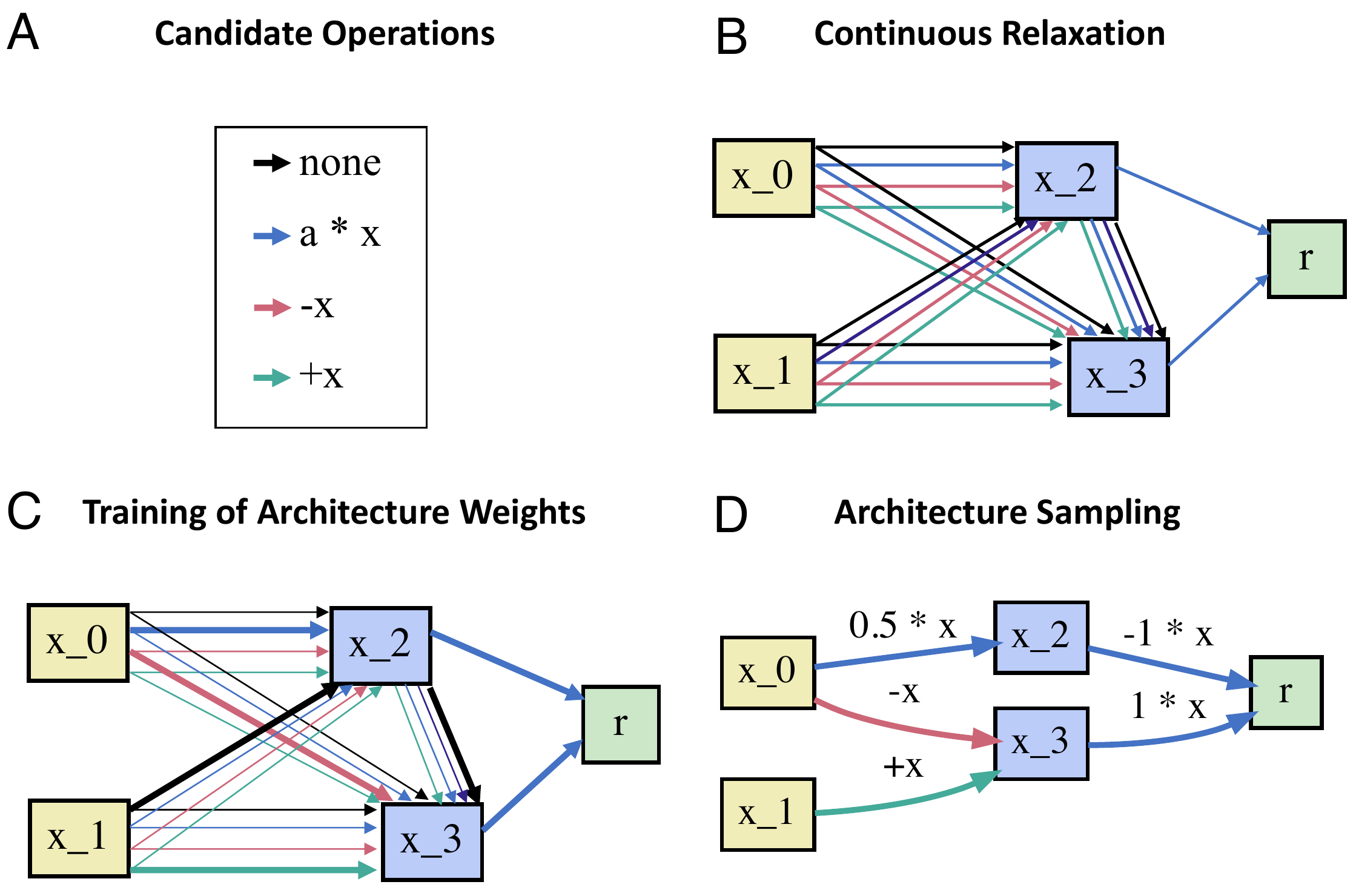}
\vspace{-0.3cm}
\caption[Learning computation graphs with DARTS.]{\textbf{Learning computation graphs with DARTS.}  The nodes and edges in a computation graph correspond to variables and functions (operations) performed on those variables, respectively. \textbf{(A)} Edges represent different candidate operations. \textbf{(B)} DARTS relaxes the search space of operations to be continuous. Each intermediate node (blue) is computed as a weighted mixture of operations. The (architectural) weight of a candidate operation in an edge represents the contribution of that operation to the mixture computation. Output nodes (green) are computed by linearly combining all intermediate nodes. \textbf{(C)} Architectural weights are trained using bi-level optimization, and used to sample the final architecture of the computation graph, as shown in \textbf{(D)}. }
\label{fig:compgraph}
\end{center}
\vspace{-0.5cm}
\end{figure}

Regular DARTS treats the architecture of a neural network as a directed acyclic computation graph (DAG), containing $N$ nodes in sequential order (Figure \ref{fig:compgraph}). Each node $x_i$ corresponds to a latent representation of the input space. Each directed edge $e_{i, j}$ is associated with some operation  $o_{i,j}$ that transforms the representation of the preceding node $i$, and feeds it to node $j$. Each intermediate node is computed by integrating over its transformed predecessors:

\vspace{-0.2cm}
\begin{equation}
\label{eqn:node_computation}
x_j = \sum_{i<j} o_{i,j} \left( x_{i} \right).
\end{equation}

Every output node is computed by linearly combining all intermediate nodes projecting to it. The goal of DARTS is to identify all operations $o_{i,j}$ of the DAG. Following \citeA{liu2018darts}, we define {$\mathscr{O} = \{o^1_{i,j}, o^2_{i,j}, \dots, o^M_{i,j}\}$} to be the set of $M$ candidate operations associated with edge $e_{i, j}$ where every operation $o^m_{i,j}(x_i)$ corresponds to some function applied to $x_{i}$ (e.g. linear,  exponential or logistic). DARTS relaxes the problem of searching over candidate operations by formulating the transformation associated with an edge as a mixture of all possible operations in $\mathscr{O}$ (cf. Figure \ref{fig:compgraph}A-B):

\vspace{-0.2cm}
\begin{equation}
\label{eqn:relaxation_original_darts}
\bar{o}_{i,j}(x) = \sum_{o \in \mathscr{O}} \frac{\textrm{exp}({\alpha^o_{i,j}})}{\sum_{o' \in \mathscr{O}} \textrm{exp}({\alpha^{o'}_{i,j}})} \cdot o_{i,j}(x).
\end{equation}
\vspace{-0.2cm}

where each operation is weighted by the softmax transformation of its architectural weight $\alpha^o_{i,j}$. Every edge $e_{i, j}$ is assigned a weight vector $\alpha_{i,j}$ of dimension $M$, containing the weights of all possible candidate operations for that edge. The set of all architecture weight vectors $\alpha = \{\alpha_{i,j}\}$ determines the architecture of the model. Thus, searching the architecture amounts to identifying $\alpha$. The key contribution of DARTS is that searching $\alpha$ becomes amenable to gradient descent after relaxing the search space to become continuous (Equation (\ref{eqn:relaxation_original_darts})). However, minimizing the loss function of the model $\mathscr{L}(w,\alpha)$ requires finding both $\alpha^*$ and $w^*$---the parameters of the computation graph.\footnote{This includes the parameters of each candidate operation $o^m_{i,j}$.} \citeA{liu2018darts} propose to learn $\alpha$ and $w$ simultaneously using bi-level optimization:

\vspace{-0.2cm}
\begin{equation}
\label{eqn:bi_level}
\begin{gathered}
\min_\alpha \mathscr{L}_{\textrm{val}}\left(w^*(\alpha),\alpha\right) \\
\textrm{s.t. } w^*(\alpha) = \underset{w}{\operatorname{argmin}}   \mathscr{L}_{\textrm{train}}(w, \alpha).
\end{gathered}
\end{equation}

That is, one can obtain $\alpha^*$ through gradient descent, by iterating through the following steps:

\begin{enumerate}
    \item Obtain the optimal set of weights $w^*$ for the current architecture $\alpha$ by minimizing the training loss $\mathscr{L}_{\textrm{train}}(w, \alpha)$.
    \item Update the architecture $\alpha$ (cf. Figure \ref{fig:compgraph}C) by following the gradient of the validation loss $\nabla  \mathscr{L}_{\textrm{val}}\left(w^*,\alpha\right)$.
\end{enumerate}

Once $\alpha^*$ is found, one can obtain the final architecture by replacing $\bar{o}_{i,j}$ with the operation that has the highest architectural weight, i.e. $o_{i,j}\leftarrow \textrm{argmax}_o \alpha^{*o}_{i,j}$ (Figure \ref{fig:compgraph}D).

\subsection{Fair DARTS}

One of the core premises of regular DARTS is that different candidate operations compete with one another in determining the transformation applied by an edge. This results from the softmax function in Equation (\ref{eqn:relaxation_original_darts}): Increasing the architectural weight $\alpha^o_{i,j}$ of operation $o$ suppresses the contribution of other operations $o'\neq o$. As a consequence, regular DARTS is biased to prefer operations that yield larger gradients (e.g. an exponential function) over operations with smaller gradients (e.g. a logistic function). To address this problem, \citeA{chu2020fair} propose to replace the softmax function in Equation (\ref{eqn:relaxation_original_darts}) with a sigmoid function such as the logistic function,

\begin{equation}
\label{eqn:logistic}
\bar{o}_{i,j}(x) = \sum_{o \in \mathscr{O}}  \frac{1}{1+\textrm{exp}(-{\alpha^o_{i,j}})} \cdot o_{i,j}(x).
\end{equation}

This introduces a cooperative (``fair'') mechanism for determining the transformation of an edge, allowing each operation to contribute in a manner that is independent from the architectural weights of other operations. To facilitate discrete encodings of the architecture, \citeA{chu2020fair} introduce a supplementary loss $\mathscr{L}_{0-1}$ that forces the sigmoid value of architectural weights toward one or zero:

\begin{equation}
\label{eqn:fair_loss}
\mathscr{L}_{0-1}=-w_{0-1}\frac{1}{N}\sum^N_l\left(\frac{1}{1+\textrm{exp}(-{\alpha_l})}-0.5\right)
\end{equation}

where $N$ corresponds to the total number of architectural weights and $w_{0-1}$ determines the contribution of $\mathscr{L}_{0-1}$ to the total loss. Here, we set $w_{0-1}=1$.

\subsubsection{Adapting DARTS for Autonomous Cognitive Modeling}

We adopt the framework from \citeA{liu2018darts} by representing  quantitative models of information processing as DAGs, and seek to automate the discovery of model architectures by differentiating through the space of operations in the underlying computation graph. To map computation graphs onto quantitative models of cognitive function, we separate the nodes of the computation graph into input nodes, intermediate nodes and output nodes. Every input node corresponds to a different independent variable (e.g. the brightness of a stimulus) and every output node corresponds to a different dependent variable (e.g. the probability of detecting the stimulus). Intermediate nodes represent latent variables of the model and are computed according to Equation (\ref{eqn:node_computation}), by applying an operation to every predecessor of the node and by integrating over all transformed predecessors.\footnote{Predecessors include both input and intermediate nodes.} For the simulation experiments reported below, we consider eight candidate operations which are summarized in Table \ref{table:operations}, including a ``zero'' operation to indicate the lack of a connection between nodes. Similar to \citeA{liu2018darts}, we compute every output node $r_j$ by linearly combining all intermediate nodes:

\vspace{-0.2cm}
\begin{equation}
\label{eqn:output_computation}
r_j = \sum_{i=S+1}^{K+S} v_{i,j} x_i
\end{equation}

where $v_{i,j} \in w$ is a trainable weight projecting from intermediate node $x_i$ to the output node $r_j$, $S$ corresponds to the number of input nodes and $K$ to the number of intermediate nodes. We seek to identify simple scientific models that---unlike complex neural networks---must be parsable by human researchers. To warrant interpretability of the model, we constrain all nodes to be scalar variables, i.e. $x_i, r_j\in \mathbb{R}^{1\times 1}$. 

Our goal is to identify a computation graph that can predict each dependent variable from all independent variables. Thus, we seek to minimize, for every dependent variable $j$, the discrepancy between every output of the model $r_j$ and the observed data $t_j$. This discrepancy can be formulated as a mean squared error (MSE), $\mathscr{L}_\textrm{MSE}(r_j, t_j \given w,\alpha)$, that is contingent on both the architecture $\alpha$ and the parameters in $w$. In addition, we seek to minimize the complexity of the model,

\vspace{-0.2cm}
\begin{equation}
\label{eqn:param_loss}
\mathscr{L}_{\textrm{complexity}} = \gamma \sum_i \sum_j \sum_m p(o^m_{i,j})
\end{equation}
\vspace{-0.2cm}

where $p(o^m_{i,j})$ corresponds to the complexity of a candidate operation, amounting to one plus the number of trainable parameters (see Table \ref{table:operations}), and $\gamma$ scales the degree to which complexity is penalized. Following the objective in Equation (\ref{eqn:bi_level}), we seek to minimize the total loss, $\mathscr{L}_\textrm{total}(w,\alpha)=\mathscr{L}_\textrm{MSE}(r_j, t_j \given w,\alpha) + \mathscr{L}_{\textrm{complexity}}$, by simultaneously finding $\alpha^*$ and $w^*$, using gradient descent.\footnote{Fair DARTS adds $\mathscr{L}_{0-1}$ (Equation \ref{eqn:fair_loss}) to the total loss.}

\begin{table}
\caption{Search space of candidate operations $o(x) \in \mathscr{O}$ and their complexity $p(o)$. Note that parameters $a,b \in w$ are fitted separately for every $o^m_{i,j}$.} 
\vspace{0.1cm}
\small
\begin{tabular}{|l r r |}
\hline
Description & $o(x)$  & $p(o)$ \\
\hline \hline
zero &  & 0 \\ 
addition & $+x$ & 1 \\ 
subtraction & $-x$ & 1 \\ 
multiplication & $a \cdot x$ & 2 \\ 
linear function & $a\cdot x + b$ & 3 \\ 
exponential function & $\textrm{exp}(a\cdot x + b)$ & 3 \\ 
rectified linear function & $\textrm{ReLU}(x)$ & 1 \\ 
logistic function & $\sigma_\textrm{logistic}(x)$ & 1 \\ 
\hline 
\end{tabular}
\label{table:operations}
\vspace{-0.4cm}
\end{table}

\section{Experiments and Results}

Identifying the architecture of a quantitative model is an ambitious task. Consider the challenge of constructing a DAG to explain the relationship between three independent variables and one dependent variable, with only two latent variables. Assuming a set of eight candidate operations per edge for a total number of seven edges, there are $8^{7}$ possible architectures to explore and endless ways to parameterize the chosen architecture. Adding one more latent variable to the model would expand the search space to $8^{12}$ possible architectures. DARTS offers one way of automating this search.  However, before applying DARTS to explain human data, it is worth assessing whether this method is capable of recovering computational motifs from a known ground truth. Therefore, we seek to evaluate whether DARTS can recover established quantitative models of human cognition from synthetic data.

As detailed below, we assess the performance of two variants of DARTS---regular DARTS and fair DARTS---in recovering three distinct computational motifs in cognitive psychology (see Test Cases). For each test case, we vary the number of intermediate nodes $k \in \{1, 2, 3\}$ and the complexity penalty $y \in \{0, 0.25, 0.5, 0.75, 1.0\}$ across architecture searches, and initialize each search with ten different seeds. 

When evaluating instantiations of NAS, it is important to compare their performance against baselines \cite{lindauer2020best}. In many cases, random search can yield results that are comparable to more sophisticated NAS \cite{li2020random,xie2019exploring}. Thus, we seek to compare the average performance of each search condition against random search. To enable a fair comparison, we allow random search to sample and evaluate architectures without replacement for the same amount of time it took either regular DARTS or fair DARTS (whichever took more time). Finally, we used the same training and evaluation procedure across all search methods.


\subsection{Training and Evaluation Procedure}

For each test case, we used 40\% of the generated data set to compute the training loss, and 10\% to compute the validation loss, to optimize the objective stated in Equation (\ref{eqn:bi_level}). We evaluated the performance of the architecture search on the remaining 50\% of the data set (test set). Experiment sequences for each data set were generated with SweetPea---a programming language for automating experimental design \cite{musslick2020sweetpea}.

For each test case and each search condition, we optimized the architecture according to Equation (\ref{eqn:bi_level}), using stochastic gradient descent (SGD). To identify $w^*$, we optimized $w$ for the selected training set over $500$ epochs with a cosine annealing schedule (initial learning rate $=0.025$, minimum learning rate $1 \times 10^{-2}$), momentum $0.9$ and weight decay $3 \times 10^{-4}$. Following \citeA{liu2018darts}, we initialize architecture variables to be zero. For a given $w^*$, we optimized $\alpha$ for the validation set over $300$ epochs using Adam \cite{kingma2014adam}, with initial learning rate $3 \times 10^{-3}$, momentum $\beta = (0.5, 0.999)$ and weight decay $1 \times 10^{-4}$. 

After training $w$ and $\alpha$, we sampled the final architecture by selecting operations with the highest architectural weights. Finally, we trained 5 random initializations of each sampled architecture on the training set for 1000 epochs using SGD with a cosine annealing schedule (initial learning rate $=0.025$, minimum learning rate $1 \times 10^{-3}$). All parameters were selected based on recoveries of out-of-sample test cases. We used the same parameters across all search methods (regular DARTS, fair DARTS and random search). All experiments were run on a 4 rack Intel cluster computer (2.5 GHz Ivybridge; 20 cores per node); each search condition was performed on a single node, allowing for 8GB memory.




\subsection{Test Cases}
All test cases are summarized in Table \ref{table:test_cases}. Here, we report the results for three different psychological models as test cases for DARTS. While these models appear fairly simple, they are based on common computational motifs in cognitive psychology, and serve as a proof of concept for uncovering potential weaknesses of DARTS.  Below, we describe each computational model in greater detail. 


\begin{table}
\caption{Summary of test cases, stating the reference to the respective equation (Eqn.), the number of independent variables (IVs), the number of dependent variables (DVs), the number of free parameters ($|\Theta|$), as well as distinct operations ($o^*$).} 
\vspace{0.1cm}
\small
\begin{tabular}{|l l l l l l|}
\hline
Test Case & Eqn. & IVs & DVs & $|\Theta|$ & o$^*$\\
\hline \hline
Weber's Law & (\ref{eqn:weber_model}) & 2 & 1 & 1 & subtraction \\ 
Exp. Learning & (\ref{eqn:exp_learning}) & 3 & 1 & 1 & exponential \\ 
LCA & (\ref{eqn:LCA_parameterized}) & 3 & 1 & 3 & rectified linear \\ 
\hline 
\end{tabular}
\label{table:test_cases}
\vspace{-0.4cm}
\end{table}

\subsubsection{Case 1: Weber's Law}

Weber's law is a quantitative hypothesis from psychophysics relating the difference in intensity of two stimuli (e.g. their brightness) to the probability that a participant can detect the difference. Here, we adopt the formal description of Weber's law with $c=1$ from Equation (\ref{eqn:weber_model}) (see Quantitative Models as Computation Graphs for a detailed description). We consider the two stimulus intensities, $I_0$ and $I_1$ as the independent variables of the model, and $P(detected)$ as the dependent variable. The generated data set (for computing $\mathscr{L}_\textrm{val}$, $\mathscr{L}_\textrm{train}$ and $\mathscr{L}_\textrm{test}$) is synthesized based on 20 evenly spaced samples from the interval $[0, 5]$ for $I_{1,2}$ and by computing $P(detected)$ for all valid crossings between $I_0$ and $I_1$, with $I_0\leq I_1$. Since we seek to explain a single probability, we apply a sigmoid function to the output of each generated computation graph. 

\subsubsection{Case 2: Exponential Learning}

The exponential learning equation is one of the standard equations to explain the improvement on a task with practice \cite{thurstone1919learning, heathcote2000power}. It explains the performance on a task $P_n$ as follows:

\vspace{-0.4cm}
\begin{equation}
\label{eqn:exp_learning}
P_n = P_\infty - (P_\infty - P_0) \cdot e^{-\epsilon \cdot t}
\end{equation}

where $t$ corresponds to the number of practice trials, $\epsilon$ is a learning rate, $P_0$ corresponds to the initial performance on a task and $P_\infty$ to the final performance for $t \rightarrow \infty$. We treat $t$, $P_0$ and $P_\infty$ as independent variables of the model and $P_n$ as a real-valued dependent variable. To avoid numerical instabilities based on large inputs, we constrain $0 \leq t \leq 1$, $0 \leq P_0 \leq 0.4$ and $0.5 \leq P_\infty \leq 1$ and set $\epsilon=5$. We generate the synthesized data set by drawing eight evenly-spaced samples for each independent variable, generating a full crossing between these samples, and by computing $P_n$ for each condition. 
The purpose of this test case is to highlight a potential weakness of DARTS: Intermediate nodes cannot represent non-linear interactions between input variables---as it is the case in Equation (\ref{eqn:exp_learning})---due to the additive integration of their inputs. Thus, DARTS must identify alternative expressions to approximate Equation (\ref{eqn:exp_learning}).

\subsubsection{Case 3: Leaky Competing Accumulator}

To model the dynamics of perceptual decision making, \citeA{usher2001time} introduced the leaky, competing accumulator (LCA). Every unit $x_i$ of the model represents a different choice in a decision making task. The activity of these units is used to determine the selected choice of an agent. The activity dynamics are determined by the non-linear equation (without consideration of noise):

\vspace{-0.4cm}
\begin{equation}
\label{eqn:LCA_general}
dx_i = [\rho_i - \lambda x_i + \mu f(x_i)  - \beta \sum_{j \neq i} f(x_j)] \frac{dt}{\tau}
\end{equation}
\vspace{-0.3cm}

where $\rho_i$ is an external input provided to unit $x_i$, $\lambda$ is the decay rate of $x_i$, $\mu$ is the recurrent excitation weight of $x_i$, $\beta$ is the inhibition weight between units, $\tau$ is a rate constant and $f(x_i)$ is a rectified linear activation function. Here, we seek to recover the dynamics of an LCA with three units, using the following (typical) parameterization: $\lambda = 0.4$, $\mu = 0.2$, $\beta = 0.2$ and $\tau=1$. In addition, we assume that all units receive no external input $\rho_i=0$. This results in the simplified equation:

\vspace{-0.2cm}
\begin{equation}
\label{eqn:LCA_parameterized}
dx_i = [- 0.4 x_i + 0.2 f(x_i)  - 0.2 \sum_{j \neq i} f(x_j)]dt
\end{equation}
\vspace{-0.3cm}

We treat units $x_1,x_2,x_3$ as independent variables ($-1 \leq x_i \leq 1$) and $dx_1$ as the dependent variable for a given time step $dt$. We generate data from the model by drawing eight evenly-spaced samples for each $x_i$, generating the full crossing between these, and computing $dx_1$ for each condition.

\subsection{Results}

\begin{figure}
\begin{center}
\includegraphics[scale = 0.48]{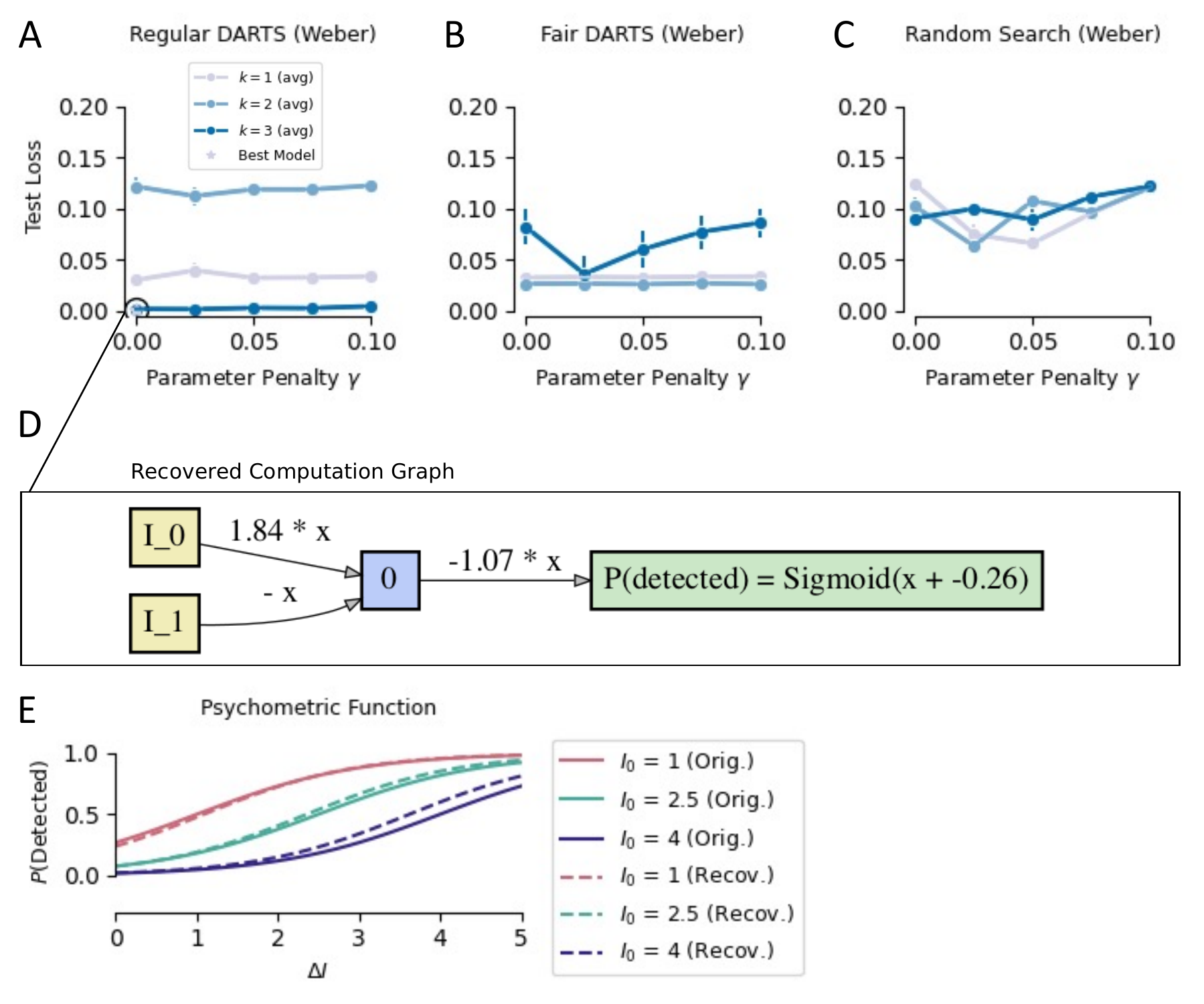}
\vspace{-0.7cm}
\caption[Architecture search results for Weber's law.]{\textbf{Architecture search results for Weber's law.}\\ \textbf{(A, B, C)} The mean test loss as a function of the number of intermediate nodes ($k$) and penalty on model complexity ($\gamma$) for architectures obtained through (A) regular DARTS, (B) fair DARTS and (C) random search. Vertical bars indicate the standard error of the mean (SEM) across seeds. The star designates the test loss of the best-fitting architecture obtained through regular DARTS, depicted in \textbf{(D)}. \textbf{(E)} Psychometric function for different baseline intensities, generated by the original model and the recovered architecture shown in (D).}
\label{fig:Weber_results}
\end{center}
\vspace{-0.5cm}
\end{figure}

Figures \ref{fig:Weber_results}, \ref{fig:exp_lrn_results} and \ref{fig:LCA_results} summarize the search results for each test case. The ground truth in each test case is generally best recovered with regular DARTS, using 3 intermediate nodes and no parameter penalty although the best fitting architecture may result from different parameters.\footnote{Note that we expect no relationship between $\gamma$ and the validation loss for random search, as random search is unaffected by $\gamma$.} Both regular and fair DARTS can achieve higher performance than random search, at least for $k=3$. Below, we examine the best-fitting architectures of regular DARTS for each test case---determined by the lowest validation loss---which are generally capable of recovering distinct operations used by the data generating model. 

\subsubsection{Case 1: Weber's Law}

The best fitting architecture ($k=1$) for Weber's law (Figure \ref{fig:Weber_results}D) can be summarized as follows:

\vspace{-0.5cm}
\begin{equation}
\label{eqn:weber_result}
P(\textrm{detected}) = \sigma_\textrm{logistic}(1.07 \cdot I_1 - 1.97 \cdot I_0 - 0.26)
\end{equation}
\vspace{-0.5cm}

and resembles a simplification of the ground truth model in Equation (\ref{eqn:weber_model}):  $\sigma_\textrm{logistic}(I_1 - 2 \cdot I_0))$, recovering the computational motif of the difference between the two input variables, as well as the role of $I_0$ as a bias term. The architecture can also reproduce psychometric functions generated by the original model (Figure \ref{fig:Weber_results}E). However, the recovery of Weber's law should be merely considered a sanity check given that the data generating model could be recovered with much simpler methods, such as logistic regression. This is reflected in the decent performance of random search.

\subsubsection{Case 2: Exponential Learning}

\begin{figure}
\begin{center}
\includegraphics[scale = 0.48]{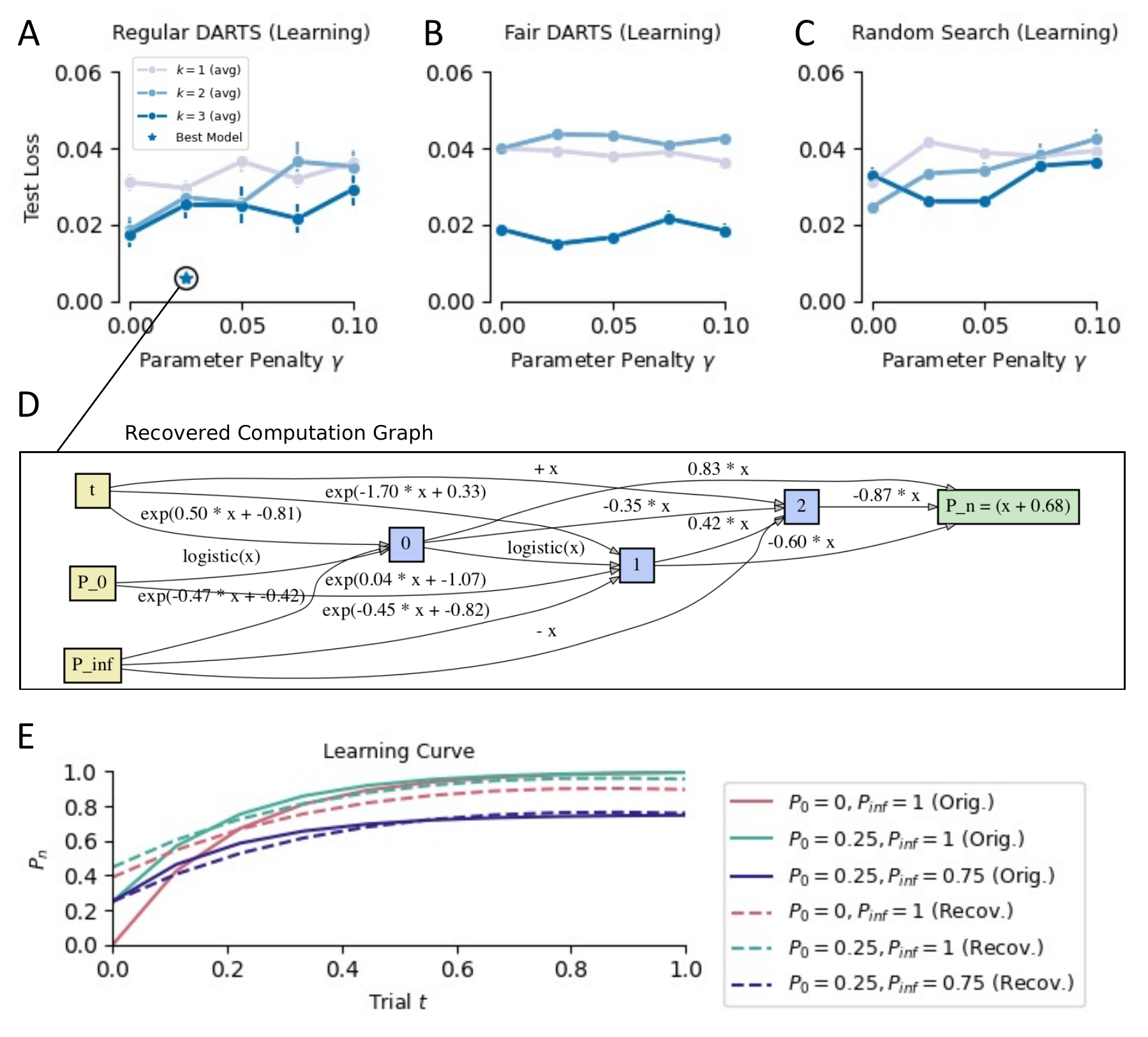}
\vspace{-0.7cm}
\caption[Architecture search results for exponential learning.]{\textbf{Architecture search results for exponential learning.} \textbf{(A, B, C)} The mean test loss as a function of the number of intermediate nodes ($k$) and penalty on model complexity ($\gamma$) for architectures obtained through (A) regular DARTS, (B), fair DARTS and (C) random search. Vertical bars indicate the SEM across seeds. The star designates the test loss of the best-fitting architecture obtained through regular DARTS, shown in \textbf{(D)}. \textbf{(E)} The learning curves generated by the original model and the recovered architecture in (D).}
\label{fig:exp_lrn_results}
\end{center}
\vspace{-0.5cm}
\end{figure}

One of the core features of this test case is the exponential relationship between task performance and the number of trials. Note that we do not expect DARTS to fully recover Equation (\ref{eqn:exp_learning}) as it is---by design---incapable of representing the non-linear interaction of $(P_\infty-P_0)$ and $e^{-\epsilon \cdot t}$. Nevertheless, regular DARTS recovers the exponential relationship between the number of trials $t$ and performance $P_n$ for $k=3$ (Figure \ref{fig:exp_lrn_results}D). However, the best-fitting architecture relies on a number of other transformations to compute $P_n$ based on its independent variables, and fails to fully recover learning curves of the original model (Figure \ref{fig:exp_lrn_results}E). In the General Discussion, we examine ways of mitigating this issue.




\subsubsection{Case 3: Leaky Competing Accumulator}

\begin{figure}
\begin{center}
\includegraphics[scale = 0.48]{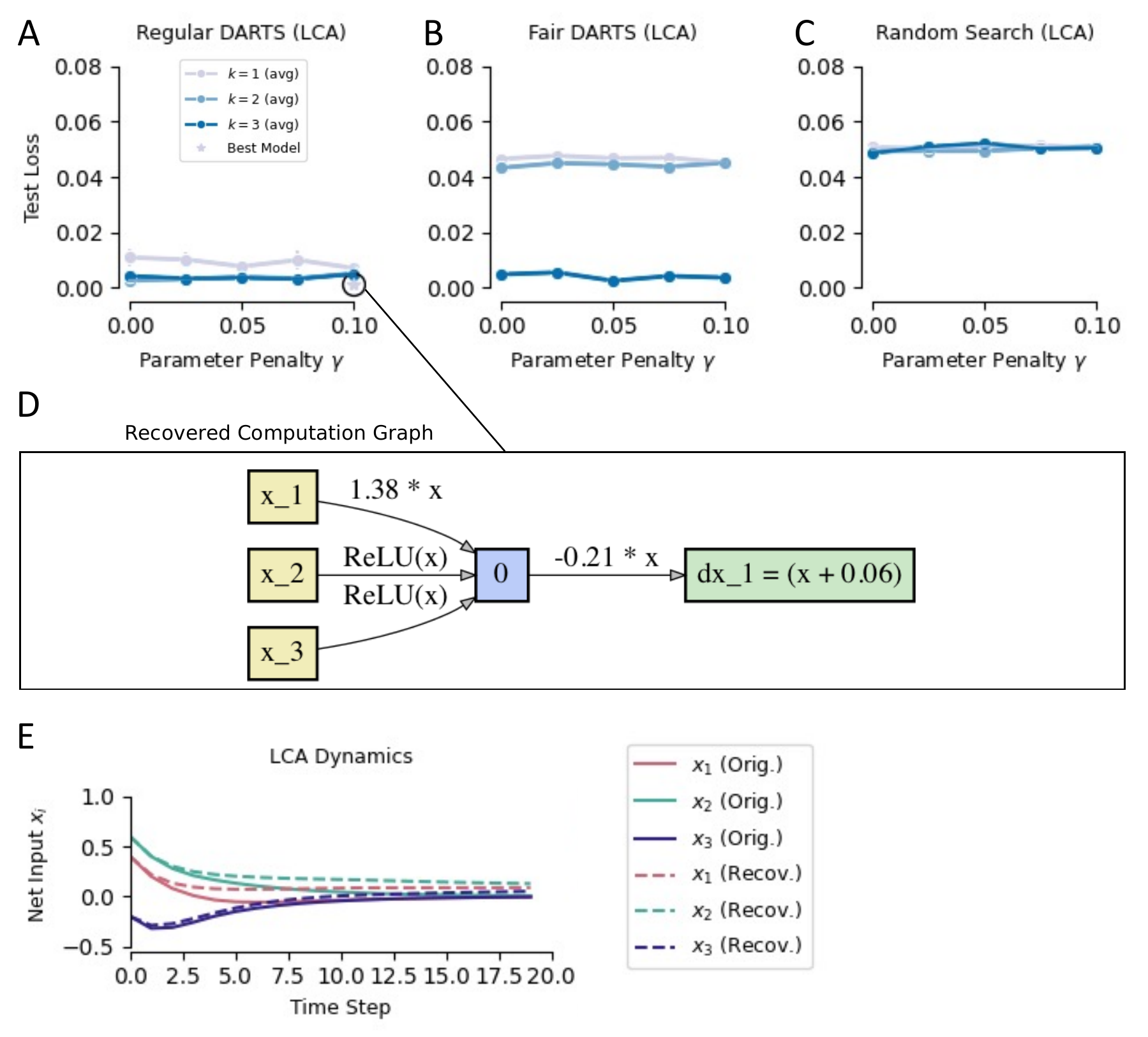}
\vspace{-0.7cm}
\caption[Architecture search results for LCA.]{\textbf{Architecture search results for LCA.} \textbf{(A, B, C)} The mean test loss as a function of the number of intermediate nodes ($k$) and penalty on model complexity ($\gamma$) for architectures obtained through (A) regular DARTS, (B) fair DARTS and (C) random search. Vertical bars indicate the SEM across seeds. The star designates the test loss of the best-fitting architecture for regular DARTS ($k=1$), depicted in \textbf{(D)}. \textbf{(E)} Dynamics of each decision unit simulated with the original model and the best architecture shown in (D), using the same initial condition at $t=0$.}
\label{fig:LCA_results}
\end{center}
\vspace{-0.5cm}
\end{figure}

The best-fitting architecture, here shown for $k=1$ (Figure \ref{fig:LCA_results}D), bears remarkable resemblance to the original model (cf. Equation (\ref{eqn:LCA_parameterized})),

\vspace{-0.3cm}
\begin{equation}
\label{eqn:LCA_result}
dx_i = [0.06 - 0.29 \cdot x_1 - 0.21 \sum_{j \neq i} \textrm{ReLU}(x_i)]dt
\end{equation}
\vspace{-0.4cm}

in that it recovers the rectified linear activation function imposed on the two units competing with $x_1$, as well as the corresponding inhibitory weight $0.21 \approx \beta = 0.2$. Yet, the recovered model misses to apply this function to unit $x_i$. However, the latter is not surprising given that the LCA has been reported to not be fully recoverable, partly because its parameters trade off against each other \cite{miletic2017parameter}.  The generated dynamics are nevertheless capable of approximating the behavior of the original model (Figure \ref{fig:LCA_results}E).



\section{General Discussion and Conclusion}


Empirical scientists are challenged with integrating an increasingly large number of experimental phenomena into quantitative models of cognitive function. In this article, we introduced and evaluated a method for recovering quantitative models of cognition using DARTS. The proposed method treats quantitative models as DAGs, and leverages continuous relaxation of the architectural search space to identify candidate models using gradient descent. We evaluated the performance of two variants of this method, regular DARTS \cite{liu2018darts} and fair DARTS \cite{chu2020fair}, based on their ability to recover three different quantitative models of human cognition from synthetic data. Our results show that these implementations of DARTS have an advantage over random search, and are capable of recovering computational motifs from quantitative models of human information processing, such as the difference operation in Weber's law or the rectified linear activation function in the LCA. While the initial results reported here seem promising, there are a number of limitations worth addressing in future work.


All limitations of DARTS pertain to its assumptions, most of which limit the scope of discoverable models. First, not all quantitative models can be represented as a DAG, such as ones that require independent variables to be combined in a multiplicative fashion (see Test Case 2). Solving this problem may require expanding the search space to include different integration functions performed on every node.\footnote{Another solution would be to linearize the data or to operate in logarithmic space. However, the former might hamper interpretability for models relying on simple non-linear functions, and the latter may be inconvenient if the ground truth cannot be easily represented in logarithmic space.} Symbolic regression algorithms provide another solution to this problem, by recursively identifying modularity of the underlying computation graph, such as multiplicative separability or simple symmetry \cite{udrescu2020ai}. Second, some operations may have an unfair advantage over others when trained via gradient descent, e.g. if their gradients are larger. This problem can be circumvented with non-gradient based architecture search algorithms, such as evolutionary algorithms or reinforcement learning. Finally, the performance of DARTS is contingent on a number of training and evaluation parameters, as is the case for other NAS algorithms. Future work is needed to evaluate DARTS for a larger space of parameters, in addition to the number of intermediate nodes and the penalty on model complexity as explored in this study. However, despite all these limitations, DARTS may provide a first step toward automating the construction of complex quantitative models based on interpretable linear and non-linear expressions, including connectionist models of cognition \cite{mcclelland1986parallel,rogers2004semantic,MusslickCohen2020}.   


In this study, we consider a small number of test cases to evaluate the performance of DARTS. While these test cases present useful proofs of concept, we encourage the rigorous evaluation of this method based on more complex quantitative models of cognitive function. To enable such explorations, we provide open access to a documented implementation of the evaluation pipeline described in this article (\url{www.empiricalresearch.ai}). This pipeline is part of a Python toolbox for autonomous empirical research, and allows for the user-friendly integration and evaluation of other search methods and test cases. As such, the repository includes additional test cases (e.g. models of controlled processing) that we could not include in this article due to space constraints. We invite interested researchers to evaluate DARTS based on other computational models, and to utilize this method for the automated discovery of quantitative models of human information processing.




\bibliographystyle{apacite}

\setlength{\bibleftmargin}{.125in}
\setlength{\bibindent}{-\bibleftmargin}

\bibliography{references}

\end{document}